# Kinematic analysis of a class of analytic planar 3-RPR parallel manipulators

Philippe Wenger and Damien Chablat

**Abstract:** A class of analytic planar 3-RPR manipulators is analyzed in this paper. These manipulators have congruent base and moving platforms and the moving platform is rotated of 180 deg about an axis in the plane. The forward kinematics is reduced to the solution of a $3^{rd}$-degree polynomial and a quadratic equation in sequence. The singularities are calculated and plotted in the joint space. The second-order singularities (cups points), which play an important role in non-singular change of assembly-mode motions, are also analyzed.

## Introduction

Solving the forward kinematic problem of a parallel manipulator often leads to complex equations and non analytic solutions, even when considering planar 3-DOF parallel manipulators [1]. For these planar manipulators, Hunt showed that the forward kinematics admits at most 6 solutions [2] and several authors [3, 4] have shown independently that their forward kinematics can be reduced as the solution of a characteristic polynomial of degree 6. Conditions under which the degree of this characteristic polynomial decreases were investigated in [5, 6]. Four distinct cases were found, all defining analytic manipulators, namely, (i) manipulators for which two of the joints coincide (ii) manipulators with similar aligned platforms (iii) manipulators with non-similar aligned platforms and, (iv) manipulators with similar triangular platforms. In cases (i), (ii) and (iv) the forward kinematics was shown to reduce to the solution of two quadratic equations in cascade while in case (iii) it was shown to reduce to a cubic and a quadratic equation in sequence. More recently, a new class of analytic manipulators was discovered [7]. These manipulators have congruent base and moving platforms and the moving platform is rotated of 180 deg about an axis in the plane. Contrary to the analytic manipulators with similar base and moving platform that are singular in the full (x, y) plane for $\varphi=0$, the recently defined analytic ones do not have this non desirable feature and, thus, might find industrial applications. Like in case (iii), their for-ward kinematics was shown to reduce to the solution of a cubic and a quadratic equation in sequence. This paper investigates the kinematics of these analytic manipulators more in detail, namely, (*i*) the number of assembly-modes and their repartition in the joint space, (*ii*) the singularity surfaces in the workspace and in the joint space, and (*iii*) the determination and the plot of the second-order singularity curves (cusps) in the joint space.

## Description of the analytic manipulator

Figure 1a shows a general 3-RPR manipulator, constructed by connecting a triangular moving platform to a base with three RPR legs. The actuated joint variables are the three link lengths $\rho_1$, $\rho_2$ and $\rho_3$. The output variables are the position coordinates ($x$, $y$) of the operation point $P$ chosen as the attachment point of link 1 to the platform, and the orientation $\varphi$ of the platform. A reference frame is centred at $A_1$ with the $x$–axis passing through $A_2$. Notation used to define the geometric parameters of the manipulator is shown in Fig. 1a.

The forward kinematics of general 3-RPR manipulators, as proposed in [3], is first briefly recalled here. The constraint equations are first set as follows:

$$\rho_1^2 = x^2 + y^2 \tag{1}$$

$$\rho_2^2 = \left(x + l_2 \cos(\varphi) - c_2\right)^2 + \left(y + l_2 \sin(\varphi)\right)^2 \tag{2}$$

$$\rho_3^2 = \left(x + l_3 \cos(\phi + \beta) - c_3\right)^2 + \left(y + l_3 \sin(\phi + \beta) - d_3\right)^2 \tag{3}$$

A system of two linear equations in $x$ and $y$ is first derived by subtracting Eq. (1) from Eqs. (2) and (3), thus obtaining a system of the form

$$Rx + Sy + Q = 0 \tag{4}$$

$$Ux + Vy + W = 0 \tag{5}$$

where the expressions of $R$, $V$, $S$, $Q$ and $W$ can be found in [3] and [7].

If $RV-SU \neq 0$, $x$ and $y$ can be solved from Eqs. (4, 5). The 6$^{\text{th}}$-degree characteristic polynomial is then obtained upon substituting the expressions of $x$ and $y$ into Eq. (1). When $RV-SU$ is equal to zero, however, the inverse kinematics cannot be established using the same procedure. The new analytic manipulators were found out in [7] by writing conditions under which $RV-SU$ is always equal to zero (i.e. for any input joint values). These conditions are [7]:

$$\begin{cases} l_2 = c_2 \\ \sin(\beta) = -d_3/l_3 \\ \cos(\beta) = c_3/l_3 \end{cases} \tag{6}$$

They means that the base and platform triangles are congruent and, because $\sin(\beta)$ is negative, the platform triangle is rotated of 180 deg about an axis in the plane. Figure 1(b) shows an instance of an analytic manipulator satisfying the above conditions.

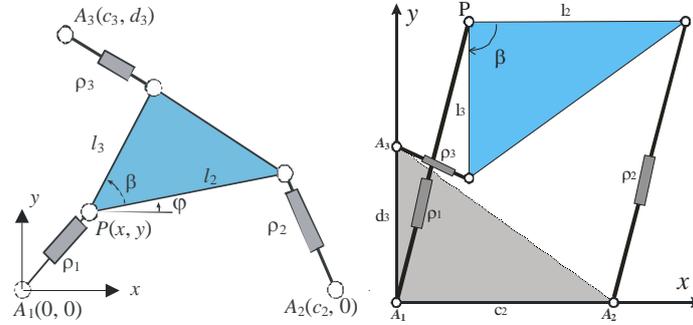

Figure 1: General 3-RPR parallel manipulator (a) and analytic (b)

## Forward kinematics and number of solutions

To obtain this polynomial, one must write that, in addition to $RV-SU=0$, $SW-VQ=0$, or, equivalently, $RW-UQ=0$. Substituting in this equation $t = \tan(\varphi/2)$ and the conditions (6) yields the following $3^{rd}$-degree characteristic polynomial for the analytic manipulators at hand (see [7] for details):

$$(c_3(\rho_1^2 - \rho_2^2 + 4c_2^2 - 4c_3c_2) + c_2(\rho_3^2 - \rho_1^2))t^3 + d_3(8c_3c_2 - 4c_2^2 + \rho_2^2 - \rho_1^2)t^2 \\ +(c_3(\rho_1^2 - \rho_2^2) + \rho_3^2 c_2 - 4d_3^2 c_2 - \rho_1^2 c_2)t + d_3(\rho_2^2 - \rho_1^2) = 0 \quad (7)$$

Once $t$ is determined from (7), rather than calculating $x$ and $y$ from (1), (4) and (5), it is more convenient to substitute $x=\rho_1\cos(\tau)$ and $y=\rho_1\sin(\tau)$ in equations (4) and (5) (where $\tau$ is the angle between $A_1P$ and the $x$ axis) and to solve one of these two equations for $\tau$, which yields up to two values for $\tau$ and for $(x, y)$.

The number of solutions of a general 3-R<u>P</u>R manipulator is fully determined by the number of real solutions of its $6^{th}$-degree characteristic polynomial. Out of singularities, this number is 0, 2, 4 or 6. For our analytic manipulators, the number of solutions is not uniquely determined by the characteristic polynomial since a quadratic equation must be solved in sequence, which may not have real solutions. For our analytic manipulator, the number of direct kinematic solutions might be 6, 4 or 2 (out of singularities), depending on the following situations:

- 6 solutions whenever (7) has 3 real roots and the quadratic equation admits 2 real roots for each root of (7);
- 4 solutions whenever (7) has three real roots and for two of these roots, the quadratic equation has two real roots and it has no real roots for the remaining root of (7);
- 2 solutions whenever (7) has only one real root and for this root the quadratic equation has 2 real roots.

In the following, we analyze the singularity surfaces of our analytic manipulators in the workspace and in the joint space.

44

## Singularity surfaces

In the literature, the singularity surfaces are generally derived in the workspace. For a general 3-RPR manipulator, these surfaces are of degree 2 in *x* and *y* and of degree 4 in *t* [8, 9]. We now derive the equation of the singularity surfaces in the workspace for the analytic manipulators under study, by substituting the conditions (1-3) into the general singularity equation. The equation can be shown to factor as follows:

$f1:\quad x+ty=0$

$f2:\quad \left(-2c_3d_3 + xd_3\right)t^3 +$
$\quad\quad \left(-2c_3^2 + 2c_2c_3 + 2d_3^2 - yd_3\right)t^2 + \left(2c_3d_3 - 2c_2d_3 + xd_3\right)t - d_3y = 0$

It is interesting to note that the first factor does not depend on the geometric parameters of the manipulator. It is linear in *x*, *y* and *t*. The second factor is also linear in *x* and *y* but it is of third order in *t*. Clearly, the singularities are much simpler than for general manipulators. Figure 2 shows the plots of the singularity surfaces for the analytic manipulator defined by $c_2 = l_2 = 1$, $c_3 = 0$, $d_3 = 1$, $l_3 = 1$ and $\beta = -\pi/2$. The surface displayed in red is associated with the first factor (the one that is independent of the manipulator geometry).

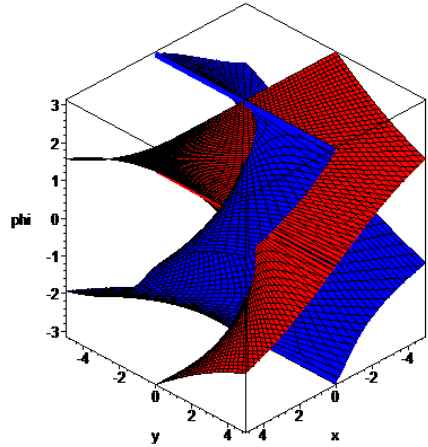

Fig. 2 Singularity surfaces in the workspace for the analytic manipulator defined by $c_2 = l_2 = 1$, $c_3 = 0$, $d_3 = 1$, $l_3 = 1$ and $\beta = -\pi/2$.

Keeping in mind that the upper and lower sides of the joint space should be "glued", it is apparent that this manipulator has two aspects (singularity-free domains [10]). Since it may have up to 6 assembly-modes, there will be two assembly-modes in one single aspect;, namely, this manipulator can change its assembly-mode without crossing a singularity if the actuated joints are unlimited. In fact, we will show in the last section that this feature is true for all analytic manipulators satisfying conditions (6).

### Singularity surfaces in the joint space

As shown in [11], it may be useful to determine the singularity locus in the joint space as well as the second-order singularity curves (cusp points) as it makes it possible to analyse non-singular assembly-mode changing motions [12-15]. Deriving the singularity surfaces in the joint space, however, is much more difficult as the algebra involved becomes extremely complex for general robots. This is why the singularity surfaces were determined only numerically in the past. We show here that for the analytic manipulators considered in this paper, the derivation of the singularity surfaces in the joint space is more tractable.

The first singularity surface is defined by the characteristic polynomial (7) and its derivative with respect to $t$. Eliminating $t$ from these two equations yields a 8$^{th}$-degree polynomial in $\rho_1$, $\rho_2$ and $\rho_3$. Because its expression is rather long, it is not reported here but it can be found at http://www.irccyn.ec-nantes.fr/~chablat/3RPR.html. The remaining singularity surfaces are defined by the quadratic equation in $x$ used to calculate the position $(x, y)$ once $t$ is found by the characteristic polynomial, and its derivative with respect to $x$. Eliminating $x$ from the quadratic equation and its derivative yields an equation in $\rho_1$, $\rho_2$, $\rho_3$ and $t$. The singularity surfaces in the joint space are then obtained by eliminating $t$ from the aforementioned equation and the characteristic polynomial (7). We then obtain a factored expression of the form $KQ_1Q_2Q_3Q_4=0$, where the $Q_i$'s are quadratics in $\rho_1$, $\rho_2$, $\rho_3$ and $K$ is a term that never vanishes. Note that we used the *resultant* function of *Maple* for all eliminations.

Finally, we come up with five independent singularity surfaces. The first one, which is the most complex, is the only surface that may contain second-order singularity points (cusps). This is because the remaining four surfaces come from a quadratic equation that cannot generate triple roots.

As an illustrative example, we now plot the singularity surfaces for the analytic manipulator analyzed in fig. 2. After assigning $c_2 = l_2 = 1$, $c_3 = 0$, $d_3 = 1$, $l_3 = 1$ and $\beta = -\pi/2$, the equation of the first singularity surface $S$ takes on the following expression:

$$S: \begin{aligned} &\rho_2^8 - 4\left(\rho_1^2 + 3\right)\rho_2^6 + \left(2\rho_3^4 + (20 - 4\rho_1^2)\rho_3^2 + 16\rho_1^2 + 44 + 8\rho_1^2(\rho_1^2 + 2)\right)\rho_2^4 \\ &+ \left((20 - 4\rho_1^2)\rho_3^4 + (8\rho_1^4 - 80\rho_1^2 - 88)\rho_3^2 - 8\rho_1^6 + 24\rho_1^4 - 32\right)\rho_2^2 \\ &\rho_3^8 - 4\left(\rho_1^2 + 3\right)\rho_3^6 + (16\rho_1^6 + 44 + 8\rho_1^4)\rho_3^4 + (24\rho_1^4 - 8\rho_1^6 - 32)\rho_3^2 \\ &-64 + \rho_1^8 - 16\rho_1^6 + 64\rho_1^2 = 0 \end{aligned}$$

The four quadratics take on the following expression:

$Q_1$: $\dfrac{1}{2}\rho_2^2 - 2 - \rho_1\rho_2 + \rho_1^2 + \rho_1\rho_3 + \dfrac{1}{2}\rho_3^2 = 0$

$Q_2$: $\dfrac{1}{2}\rho_2^2 - 2 + \rho_1\rho_2 + \rho_1^2 + \rho_1\rho_3 + \dfrac{1}{2}\rho_3^2 = 0$

$Q_3$: $\dfrac{1}{2}\rho_2^2 - 2 + \rho_1\rho_2 + \rho_1^2 - \rho_1\rho_3 + \dfrac{1}{2}\rho_3^2 = 0$

$Q_4$: $\dfrac{1}{2}\rho_2^2 - 2 - \rho_1\rho_2 + \rho_1^2 - \rho_1\rho_3 + \dfrac{1}{2}\rho_3^2 = 0$

Figure 3 shows the plots of the five singularity surfaces in a section ($\rho_1$, $\rho_3$) a of the joint space defined by $\rho_2$=1. The $Q_i$'s form portions of ellipses in the quadrant $\rho_1$>0 and $\rho_3$>0 (correspondence is indicated on the figure). The curve that includes cusp points is associated with *S*. As shown theoretically some years ago [8], the joint space is divided into regions (called *basic components*) where the number of assembly-modes is 2, 4 or 6. These regions are separated by the singular surfaces. Under the action of the forward kinematics, these regions give rise to several distinct regions in the workspace (called *basic regions* [8]), each one being associated with an assembly-mode. The number of solutions in each region is represented in Fig. 3 by distinct colours: green for 2 solutions, red for 6 solutions and yellow for 4 solutions.

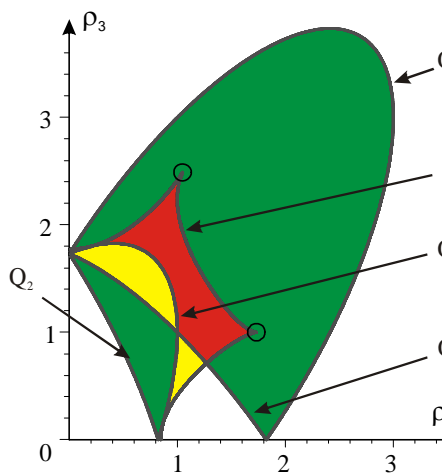

Fig. 3 Singularity surfaces in section $\rho_2$=1 of the joint space for the analytic manipulator defined by $c_2 = l_2 = 1$, $c_3 = 0$, $d_3 = 1$, $l_3 = 1$ and $\beta = -\pi/2$. The two cusp points are shown with circles.

## Second-order singularity curves (cusps)

The second-order singularities play an important role in non-singular assembly-mode changing motions: such motions are possible only if the 3-R<u>P</u>R manipulator has cusp points in sections of its singularity surfaces in the joint space [12, 14]. The 4 analytic manipulators defined in the past (cases (i), (ii), (iii) and (v) recalled in the introduction) do not have any cusp points. This is because the polynomials involved in their forward kinematics cannot have triple roots. Indeed, manipulators of cases (i), (ii) and (iv) have a characteristic polynomial of degree 2 and in case (iii) the cubic was shown to have only two real roots [5].

Instead, the analytic manipulators studied in this paper can be shown to have always cusp points and, thus, they can perform non-singular assembly mode changing motions. Let first determine the second-order singularity curves. Contrary to general 3-R<u>P</u>R manipulators that do not lend themselves easily to such calculations because of the complex algebra involved, the analytic manipulators studied in this paper are much simpler to analyze. The second-order singularities can be defined by Eq. (7) and its first and second derivatives with respect to *t*.

Combining these three equations and eliminating *t* and one of the joint variables, say $\rho_2$, makes it possible to determine the projection onto ($\rho_1$, $\rho_3$) of the second-order singularity curves. Setting $c_2=1$ without loss of generality and after eliminating spurious solutions, the resulting equation is a cubic in $R3=\rho_3^2$ shown in Eq. (8) below.

$$\begin{aligned}
\Big( & 12\,d3^2\,\rho1^2\,c3 - 15\,d3^2\,c3^3 - 6\,d3^2\,\rho1^2\,c3^2 + \frac{69}{4}d3^2\,c3^4 - \frac{27}{2}c3^5 + \frac{27}{4}c3^6 + \frac{57}{4}d3^4\,c3^2 \\
& -\frac{3}{2}d3^4\,c3 + \frac{27}{4}d3^4 - 6\,d3^4\,\rho1^2 + \frac{3}{2}d3^2\,c3^2 + \frac{27}{4}c3^4 + \frac{15}{4}d3^6 - 3\,d3^2\,\rho1^4 \Big) R3 \\
& + (3\,d3^2\,c3^2 + 3\,d3^4 + 3\,d3^2\,\rho1^2 - 6\,d3^2\,c3)\,R3^2 - d3^2\,R3^3 - \frac{27\,c3^6\,\rho1^2}{4} \\
& + 3\,c3^2\,d3^6 - \frac{57\,d3^4\,\rho1^2\,c3^2}{4} - 6\,c3\,d3^6 + 3\,d3^4\,\rho1^4 - 6\,d3^2\,c3^5 + c3^6\,d3^2 \\
& - \frac{69\,d3^2\,\rho1^2\,c3^4}{4} - \frac{3\,d3^2\,\rho1^2\,c3^2}{2} - 8\,d3^2\,c3^3 + 12\,d3^2\,c3^4 - 6\,d3^2\,\rho1^4\,c3 \\
& + 3\,d3^2\,\rho1^4\,c3^2 + 15\,d3^2\,\rho1^2\,c3^3 - \frac{27\,c3^4\,\rho1^2}{4} + d3^2\,\rho1^6 + \frac{27\,c3^5\,\rho1^2}{2} + 3\,d3^4\,c3^4 \\
& - \frac{27\,d3^4\,\rho1^2}{4} - 12\,d3^4\,c3^3 + \frac{3\,d3^4\,\rho1^2\,c3}{2} + 12\,d3^4\,c3^2 + d3^8 - \frac{15\,d3^6\,\rho1^2}{4} = 0
\end{aligned} \quad (8)$$

Now, the derivative of this cubic has two roots and one of them can be shown to be always positive. For this positive root, the left-hand side of (8) can be shown to be also positive. Since, in addition, the coefficient of $R3^3$ is negative, the left-hand side of (8) tends to $-\infty$ when $R3 \to +\infty$. This shows that the cubic has at least one positive root. Thus, the analytic manipulators under study have cusp points. In particular, the manipulator analyzed in figs 2 and 3 has two cusp points at $\rho_2=1$ (encircled in Fig. 3), located at ($\rho_1$= 1.73205080, $\rho_3$=1) and ($\rho_1$= 1.04789131, $\rho_3$=2.48920718).

## Conclusions

This paper was devoted to the kinematic analysis of a family of analytic 3-R<u>P</u>R manipulator recently discovered. Their forward kinematics can be solved with a cubic and a quadratic equation in sequence. The equation of the singularity surfaces were derived both in the workspace and in the joint space. In the workspace, the singularity equation was shown to be composed of two factors. One is linear in *t*, *x* and *y* and, surprisingly, is independent of the manipulator geometry. The other one is linear in *x* and *y* and of degree 3 in *t*. In the joint space, the singularity equation could be determined symbolically for the first time. It was show to be composed of five factors. One of them is of degree 8 in $\rho_1$, $\rho_2$ and $\rho_3$ and contains the cusps. The remaining four factors are quadratics. Finally, the second-order singularity curves were determined and it was shown that, contrary to all other analytic manipulators found in the past, the ones analyzed in this paper have always cusp point and, thus, can perform non-singular assembly-mode changing motions.